\documentclass[conference]{IEEEtran}
\IEEEoverridecommandlockouts
\usepackage{cite}
\usepackage{amsmath,amssymb,amsfonts}
\usepackage{algorithmic}
\usepackage{graphicx}
\usepackage{booktabs}
\usepackage{multirow}
\usepackage{textcomp}
\usepackage{xcolor}
\def\BibTeX{{\rm B\kern-.05em{\sc i\kern-.025em b}\kern-.08em
    T\kern-.1667em\lower.7ex\hbox{E}\kern-.125emX}}
\begin{document}

\title{Leveraging Deep Learning and Digital Twins to Improve Energy Performance of Buildings
\thanks{This study has been financially supported by the Swedish Energy Agency.}
}

\author{\IEEEauthorblockN{1\textsuperscript{st} Zhongjun Ni}
\IEEEauthorblockA{\textit{Department of Science and Technology} \\
\textit{Link\"oping University, Campus Norrk\"oping}\\
Norrk\"oping, Sweden \\
zhongjun.ni@liu.se} \\
\IEEEauthorblockN{3\textsuperscript{rd} Magnus Karlsson}
\IEEEauthorblockA{\textit{Department of Science and Technology} \\
\textit{Link\"oping University, Campus Norrk\"oping}\\
Norrk\"oping, Sweden \\
magnus.b.karlsson@liu.se}
\and
\IEEEauthorblockN{2\textsuperscript{nd} Chi Zhang}
\IEEEauthorblockA{\textit{Department of Computer Science and Engineering} \\
\textit{University of Gothenburg}\\
Gothenburg, Sweden \\
chi.zhang@gu.se} \\
\IEEEauthorblockN{4\textsuperscript{th} Shaofang Gong}
\IEEEauthorblockA{\textit{Department of Science and Technology} \\
\textit{Link\"oping University, Campus Norrk\"oping}\\
Norrk\"oping, Sweden \\
shaofang.gong@liu.se}
}

\maketitle

\begin{abstract}

Digital transformation in buildings accumulates massive operational data, which calls for smart solutions to utilize these data to improve energy performance. This study has proposed a solution, namely Deep Energy Twin, for integrating deep learning and digital twins to better understand building energy use and identify the potential for improving energy efficiency. Ontology was adopted to create parametric digital twins to provide consistency of data format across different systems in a building. Based on created digital twins and collected data, deep learning methods were used for performing data analytics to identify patterns and provide insights for energy optimization. As a demonstration, a case study was conducted in a public historic building in Norrk\"oping, Sweden, to compare the performance of state-of-the-art deep learning architectures in building energy forecasting. 

\end{abstract}

\begin{IEEEkeywords}
deep learning, digital twin, building energy forecasting
\end{IEEEkeywords}

\section{Introduction}
\label{sec:introduction}

%% P1
Digital transformation in buildings has brought considerable opportunities to optimize their energy performance by integrating various advanced information and communication technologies~\cite{liu_methodology_2021}. Among them, digital twin technology is today a powerful tool for building management. With continuously collected data~\cite{ni_link_2022}, a digital twin reflects the latest status of its physical counterpart in nearly real-time~\cite{wang_survey_2022}. In addition, more advanced data analysis applications, such as energy forecasting and predictive controls, can be developed based on the virtual model and data from meters, sensors, actuators, and control systems~\cite{burakgunay_data_2019}. Deep learning has shown great potential in data analytics~\cite{liu_anomaly_2020,zhang_social_2021}. Based on large amounts of the collected data, deep learning methods can be used to develop models for identifying patterns in operational data, such as making predictions about energy use and revealing the potential for energy optimization~\cite{ni_improving_2021}. 
%% End P1

%% P2
Previous studies have demonstrated the benefits of digital twins for the built environment, such as indoor or ambient climate monitoring~\cite{khajavi_digital_2019, rosati_air_2020} and anomaly detection for building assets~\cite{lu_digital_2020}. However, these studies lack an emphasis on the consistency of data representation of virtual models. Data collected from buildings are usually produced by various systems and methods~\cite{acierno_architectural_2017}. Even the most advanced building management system generates a bluster of data and information flows that differ among buildings, vendors, and locations~\cite{balaji_brick_2018}. Lacking a consistent data format makes it challenging to extend previous solutions to other buildings and limits the deployment of energy applications. Furthermore, the integration of deep learning and digital twins is still in its early stage. State-of-the-art deep learning architectures, e.g., temporal fusion transformer (TFT)~\cite{lim_temporal_2021}, have not been exploited in the built environment for improving energy performance. 
%% End P2

%% P3
This study aims to integrate deep learning and digital twins to better understand building energy usage. The main contributions of this work are:
\begin{itemize}
  \item A solution, namely Deep Energy Twin, was proposed for analyzing building energy use and identifying potential for energy optimization. Ontology was adopted to create parametric digital twins to provide consistency of data format across different systems in a building. Deep learning was used for data analytics.
  \item A comprehensive case study was conducted to illustrate the capacity of five deep learning methods, including long short-term memory (LSTM), temporal convolutional network (TCN), Transformer, N-HiTS, and TFT, to predict building energy consumption and measure uncertainties. 
\end{itemize}
%% End P3

\section{Related Work}
\label{sec:related_work}

This section first introduces the application of digital twins in the built environment. Then, deep learning methods for time series forecasting are reviewed.

\subsection{Digital Twins in the Built Environment}

Several studies have reported developing digital twin applications for the built environment, such as indoor or ambient climate monitoring~\cite{khajavi_digital_2019, rosati_air_2020}, anomaly detection for building assets~\cite{lu_digital_2020}, and heritage preservation~\cite{zhang_automatic_2022, ni_enabling_2022}. However, most of the studies lack an emphasis on the consistency of data representation of virtual models. They typically employ some customized data format, which makes it difficult to extend their solutions to other buildings, limits interoperability between buildings, and limits the deployment of energy applications. Only little work~\cite{ni_enabling_2022} looked into using a consistent metadata structure to represent buildings and subsystems. Nevertheless, integrating energy optimization solutions in buildings requires expertise in multiple domains~\cite{balaji_brick_2018}. For depicting such a complex system, it is preferable to use ontology to ensure accurate alignment across several domains, such as actuators, sensors, management workflows, and web resources~\cite{boje_semantic_2020}.    

An ontology is a formal statement of a conceptualization that includes the objects, concepts, and other entities presumed to exist in a given area, together with the relationships between them~\cite{gruber_principles_1995}. Several studies have attempted to tackle the challenge of creating a metadata schema across a broad range of buildings. Balaji et al.~\cite{balaji_brick_2018} proposed Brick, a standardized metadata schema for representing buildings. The schema defines a concrete ontology for sensors, subsystems, and their relationships, enabling the development of portable applications. RealEstateCore~\cite{hammar_realestatecore_2019} is another ontology for the real estate business to speed up building modeling. Both Brick and RealEstateCore allow data output in the Digital Twin Definition Language~\cite{dtdl_v2} format, which facilitates deploying applications in Microsoft Azure Cloud. Using these ontologies to create virtual models is advantageous for gathering and documenting all necessary information for further knowledge management and data analytics.

\subsection{Deep Learning for Time Series Forecasting}

Deep learning methods have emerged in recent years due to their enhanced abilities in addressing massive data, feature extraction, and modeling nonlinear processes~\cite{runge_review_2021}. Three fundamental deep learning methods for time series forecasting are recurrent neural networks (RNNs), convolutional neural networks (CNNs), and attention mechanism-based networks. 

In previous studies, RNN and its variants~\cite{hochreiter_long_1997} have been more frequently applied to building energy forecasting~\cite{kong_short-term_2019, fan_assessment_2019}. A few studies~\cite{lara-benitez_temporal_2020,wang_transformer-based_2022} have also applied TCN~\cite{bai_empirical_2018} and Transformer~\cite{vaswani_attention_2017}. However, recent deep learning methods, such as TFT~\cite{lim_temporal_2021} and N-HiTS~\cite{challu_n-hits_2022}, were rarely used. Therefore, a practical comparison of which method is more effective in building energy forecasting is lacking. In addition, previous studies mostly made point forecasting, and little work was carried out on making probabilistic forecasting.

\section{Methodology}
\label{sec:methodology}

First, the process of creating a parametric digital twin of building energy systems is presented. Then, the method for developing predictive models for a representative building energy application is described.

\subsection{Creation of Parametric Digital Twins}

\begin{figure}[!ht]
\includegraphics[width=\columnwidth]{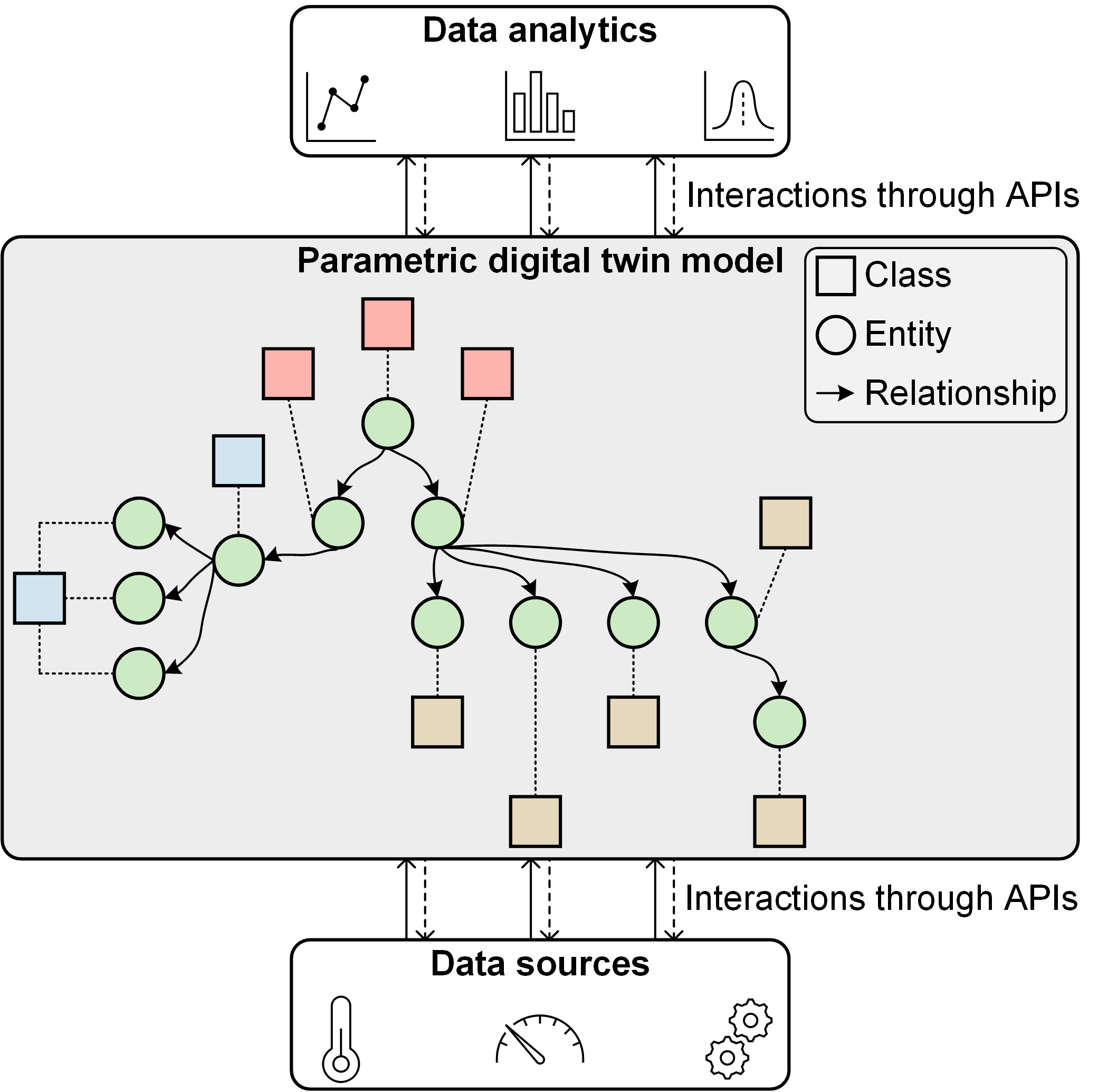}
\centering
\caption{An illustration of the parametric digital twin model. A parametric digital twin model contains essential information about physical entities and their relationships. The model also provides necessary read or write application programming interfaces (APIs) for other modules, such as updating model status from data sources and providing data access for data analytics.}\label{fig:parametric_dt}
\end{figure}

As depicted in Fig.~\ref{fig:parametric_dt}, creating a parametric digital twin of a building lies in two aspects. One is to model essential physical entities and their relationships. The other is to provide the necessary interfaces to continuously update the status of entities from various data sources and supply data access for subsequent tasks, e.g., data analytics. As a reference implementation, the Brick ontology~\cite{balaji_brick_2018} was adopted for creating the parametric digital twin model. The ontology provides a consistent data representation that converts heterogeneous energy system data to a consistent format. In Fig.~\ref{fig:parametric_dt}, square boxes represent classes, which abstract physical entities, such as \textit{Location}, \textit{Equipment}, \textit{Resource}, and \textit{Point}. Each round box represents a specific entity, which is an instance of a particular class. A class can have multiple instances of entities.

Locations refer to different spaces in buildings, such as rooms and floors. Resources are physical resources or stuff that are controlled or measured by points. Physical or virtual entities that generate time series data are called points. Typical physical points include sensors, setpoints, and equipment status. Virtual points, on the other hand, are generated by a mechanism that may operate on other time series data, such as an average floor humidity sensor. Each data point can have several relationships that connect it to other classes, such as its location or the equipment it belongs to.

\subsection{Building Energy Forecasting}

As a demonstration of data analytics, this subsection introduces one-step ahead building energy forecasting.

\subsubsection{Problem Formulation}

A specific energy use, i.e., target variable, is denoted as $y$ and $y\in \mathbb{R}_{+}$.
Predictor variables that affect energy use are denoted as $\mathbf{x}$ and $\mathbf{x} \in \mathbb{R}^{k}$.
All target and predictor variables are supposed to be observed at constant intervals over time and grouped chronologically.
At time $t$, the observed value of the target variable is denoted as ${y}_{t}$.
Similarly, observed values of predictor variables are denoted as $\mathbf{x}_{t} =[{x}_{1,t} ,{x}_{2,t} ,...,{x}_{k,t}]$.

Then, a point forecasting model takes the form
\begin{equation}
\hat{y}_{t+1} =f_{\mathbf{\theta }}(y_{t-w+1:t} ,\mathbf{x}_{t-w+1:t}), \label{eqn:point_forecast}
\end{equation}
where $\hat{y}_{t+1}$ is model forecast, $y_{t-w+1:t} = [y_{t-w+1} ,y_{t-w+2} ,...,y_{t}] $ and $\mathbf{x}_{t-w+1:t}=\{\mathbf{x}_{t-w+1} ,\mathbf{x}_{t-w+2} ,...,\mathbf{x}_{t}\}$ are observations of the target and predictor variables over a loopback window $w$, and $f_{\mathbf{\theta }}{(.)}$ is the prediction function learned by the model.

Probabilistic forecasting models are developed to generate interested quantiles directly through quantile regression. Given a set of quantiles $\mathcal{Q} \subset (0,1)$, a quantile forecasting model takes the form
\begin{equation}
\hat{y}_{t+1}^{(p)} =g_{\mathbf{\theta }}(y_{t-w+1:t} ,\mathbf{x}_{t-w+1:t}), \label{eqn:quantile_forecast}
\end{equation}
where $p \in \mathcal{Q}$, $\hat{y}_{t+1}^{(p)}$ is the model forecast for the $p$th quantile of the target variable, $y_{t-w+1:t}$ and $\mathbf{x}_{t-w+1:t}$ have the same definition as in the point forecasting model, and $g_{\mathbf{\theta }}{(.)}$ is the prediction function learned by the model.

\subsubsection{Deep Learning Methods for Comparison}

Five deep learning methods, namely LSTM, TCN, Transformer, N-HiTS, and TFT, were investigated to compare their performance in building energy forecasting.

\subsubsection{Loss Function and Evaluation Metrics}
\label{sec:loss_func_eval_metric}

Point forecasting models were trained to minimize the total squared error. Probabilistic forecasting models were trained to minimize the total quantile loss. The $p$th quantile loss~\cite{lim_temporal_2021} is calculated as 
\begin{equation}
\ell(\hat{y} ,y,p)=(1-p)(\hat{y} -y)_{+} + p(y-\hat{y})_{+}, \label{eqn:quantile_loss}
\end{equation}
where $(.)_{+}=max(0,.)$. Then, the training quantile loss $L_{q}(\mathbf{\theta})$ for a set $\mathcal{S} =\{(y_{t-w+1:t} ,\mathbf{x}_{t-w+1:t} ,y_{t+1})\}_{t=w}^{n+w-1}$ is calculated as
\begin{equation}
L_{q}(\mathbf{\theta }) =\sum _{t=w}^{n+w-1}\sum _{i=1}^{| \mathcal{Q}| } \ell \left(\hat{y}_{t+1}^{( p_{i})} ,y_{t+1} ,p_{i}\right),
\end{equation}
where $p_{i} \in \mathcal{Q}$.

The prediction accuracy of point forecasting models was evaluated by coefficient of variation of the root mean square error (CV-RMSE) and normalized mean bias error (NMBE). They are calculated by Eq.~\ref{eqn:cv-rmse} and~\ref{eqn:nmbe}~\cite{ashare_measurement_2014}.
\begin{equation}
CV\textrm{-}RMSE=\frac{\sqrt{\frac{1}{n}\sum\limits _{t=1}^{n}(\hat{y}_{t} -y_{t})^{2}}}{\overline{y}} \times 100, \label{eqn:cv-rmse}
\end{equation}
\begin{equation}
NMBE=\frac{\frac{1}{n}\sum\limits _{t=1}^{n}(\hat{y}_{t} -y_{t})}{\overline{y}} \times 100,  \label{eqn:nmbe}
\end{equation}
where $n$ denotes the size of forecast horizon, $y_{t}$ and $\hat{y}_{t}$ have the same definitions in the point forecasting model. $\overline{y}$ is the mean actual value of the target variable over the forecast horizon.

The $\rho$-risk, which normalizes quantile losses, was used for evaluating the performance of probabilistic forecasting models. $\rho$-risk at $p$th quantile is calculated by~\cite{lim_temporal_2021}
\begin{equation}
\rho\textrm{-}risk(p) =\frac{2\times \sum\limits _{t=1}^{n} \ell \left(\hat{y}_{t}^{(p)} ,y_{t} ,p\right)}{\sum\limits _{t=1}^{n} y_{t}}, \label{rho_risk}
\end{equation}
where $n$ denotes the size of forecast horizon, $\hat{y}_{t}^{(p)}$ is the predicted $p$th quantile value of a target variable at time $t$, and $\ell \left(\hat{y}_{t}^{(p)} ,y_{t} ,p\right)$ is the $p$th quantile loss calculated by Eq.~\ref{eqn:quantile_loss}. 

\section{Case Study}
\label{sec:case_study}

To verify the performance of different deep learning methods, a case study was conducted to develop predictive models for energy use of one public historic building. This section describes details of the used dataset and experimental setup.

\subsection{Dataset}
\label{sec:dataset}
\begin{figure}[!ht]
\includegraphics[width=\columnwidth]{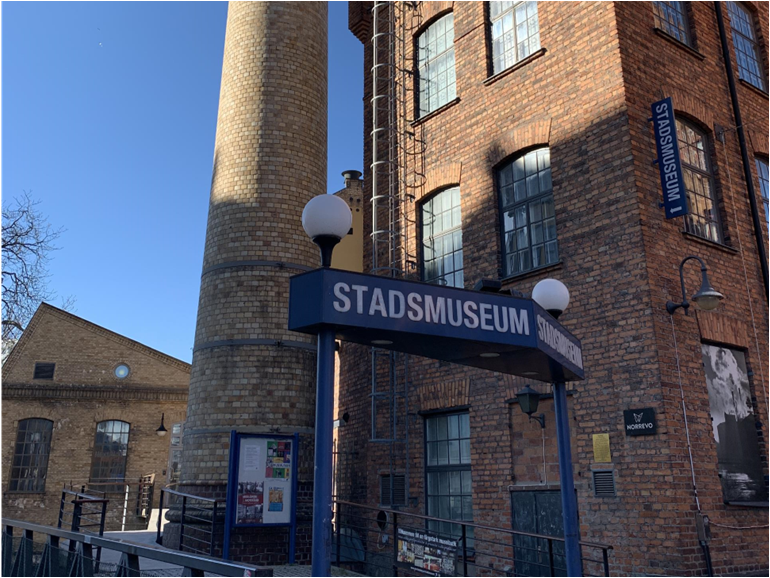}
\centering
\caption{The City Museum, Norrk\"oping, Sweden.}\label{fig:the_city_museum}
\end{figure}

The dataset includes two parts. One is the historical electricity consumption and heating load from the City Museum (see Fig.~\ref{fig:the_city_museum}) in Norrköping, Sweden. The other is the meteorological data from a weather station located $\sim$2~km away from the building. The meteorological data include dry-bulb temperature, relative humidity, dew point temperature, precipitation, air pressure, and wind speed. All data range from 01:00 on January 1, 2016 to 00:00 on January 1, 2020, with a time granularity of one hour. Hours appearing in this paper are expressed in 24-hour format and are in local time.

\begin{figure}[!ht]
\includegraphics[width=\columnwidth]{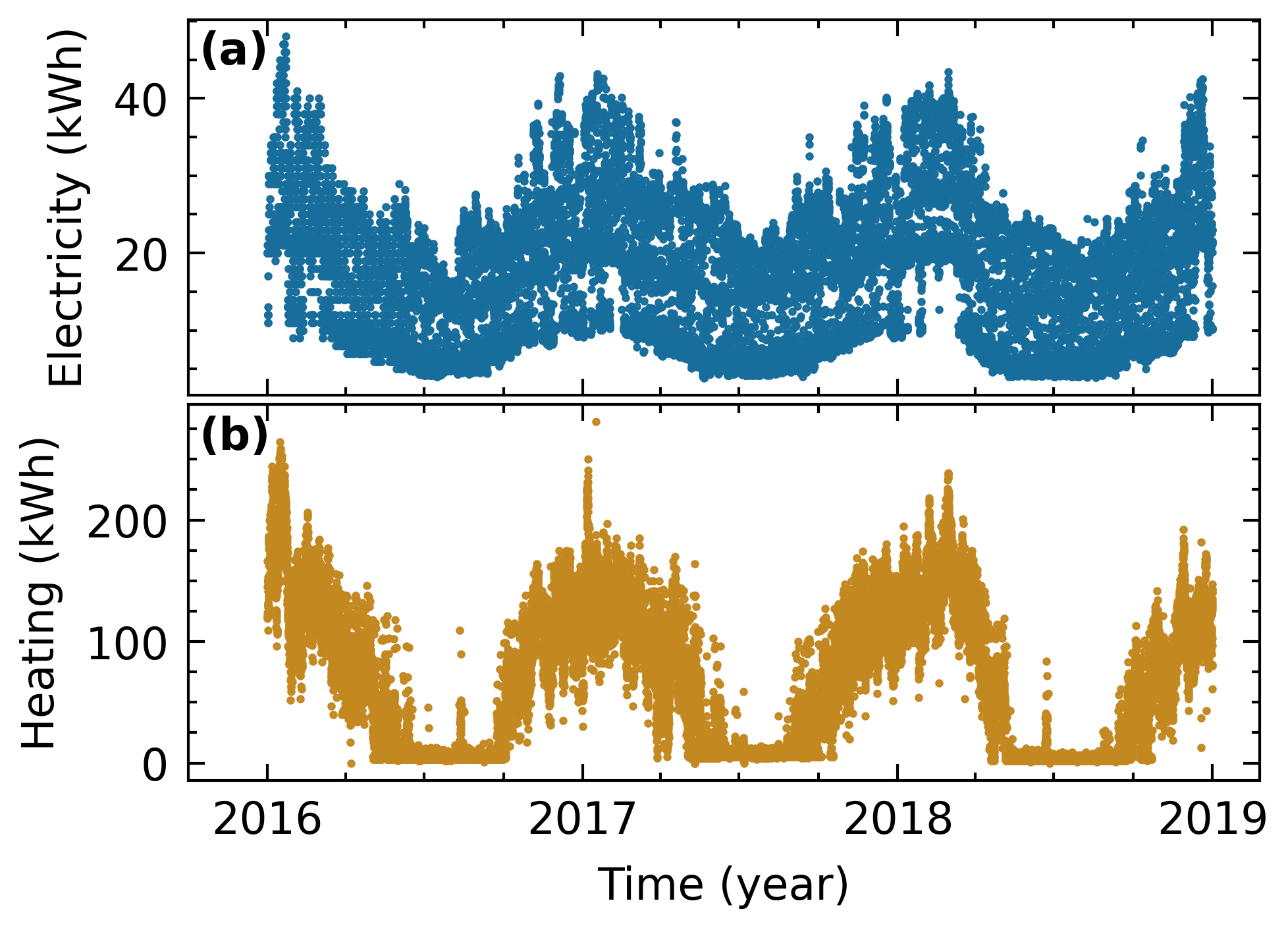}
\centering
\caption{Historical hourly (\textbf{a}) electricity consumption and (\textbf{b}) heating load of the City Museum from 01:00 on January 1, 2016 to 00:00 on January 1, 2019.}\label{fig:energy}
\end{figure}

The normal operation of the City Museum is to maintain an appropriate indoor climate for preservation of collections and human comfort of staff and visitors. In regular time, it is open six days a week, from Tuesday to Sunday. The opening time starts at 11:00. The closing time is 17:00 on Tuesdays, Wednesdays, and Fridays, 20:00 on Thursdays, and 16:00 on Saturdays and Sundays. According to the first three years of energy consumption data (see Fig.~\ref{fig:energy}), both electricity and heating have a yearly seasonality. Moreover, there is no long-term trend in both energy use.

\subsection{Data Preprocessing}
\label{sec:data_preprocessing}
Data preprocessing seeks to turn raw data into a format that models can readily handle and understand.

\subsubsection{Data Cleaning and Dataset Splitting}

First, missing values in meteorological data were linearly interpolated. Then, the dataset was partitioned into three subsets: a training set for learning model parameters, a validation set for tuning hyperparameters and avoiding overfitting, and a test set for evaluating model performance. The dataset is divided roughly according to the empirical ratio of 80:10:10, with 38 months of data from January 1, 2016 to February 28, 2019 are utilized as the training set, five months of data from March 1, 2019 to July 31, 2019 are used as the validation set, and the remaining five months of data are used as the test set.

\subsubsection{Feature Preparation}
\label{sec:feature_prep}
Four temporal features are extracted from timestamps: two cyclical and two binary variables. The cyclical variables are \textit{hour} (integer value from 0 to 23) and \textit{weekday} (integer value from 0 to 6, each value represents a day in a week, starting from Monday). The binary variables include one called \textit{is holiday} to indicate if a day is a Sweden public holiday and another called \textit{is weekend} to indicate if a day is a weekend. In addition, one feature called \textit{is open} with a binary value is added to indicate if the City Museum is open for a given hour.

\subsubsection{Data Transformation}
A min-max normalization was used to scale target variables and meteorological features to a range of $[0, 1]$. The training set was utilized to fit all min-max scalers, which were then used to transform the validation and test sets. A sine-cosine transformation was used to convert cyclical features. Binary features were not transformed.

\subsection{Experimental Setup}
\label{sec:experimental_setup}

Models were developed to forecast electricity consumption and heating load of the City Museum in the next hour. A lookback window size of 24 was used. Three baseline models were created for point forecasting. Two of them are based on the seasonal na\"ive (SN) method~\cite{hyndman_forecasting_2018}, namely SN-1 and SN-24, because electricity consumption and heating load are highly seasonal. Each forecast of the SN-1 model is set to be the value observed one hour ago. The SN-24 model seeks to use daily seasonality, and each forecast of a target variable is set to the value observed 24 hours ago. The remaining one is linear regression (LR) model.

Five deep learning models (LSTM, TCN, Transformer, N-HiTS, and TFT) were trained to make both point and probabilistic forecasts. The predefined set of quantiles is $\{0.1,0.5,0.9\}$, and we are interested in evaluating $\rho$-risk on 0.5th and 0.9th quantiles. Models were built using the Python packages PyTorch (v1.12.0), darts (v0.23.1), and scikit-learn (v1.2.1). All experiments were carried out on a computer equipped with an NVIDIA GeForce GTX 1080 graphics card. 

\section{Results and Discussion}
\label{sec:results_and_discussion}

\subsection{Quantitative Analysis}

As shown in Table~\ref{tab:point_forecast_result}, among all models, the TCN model performed best on predicting both energy use (CV-RMSE 13.90\% on electricity and CV-RMSE 8.16\% on heating). The performance of all models except for the SN-24 model on forecasting electricity has met the criterion suggested by the ASHRAE Guideline 14-2014~\cite{ashare_measurement_2014} (30\% for CV-RMSE and $\pm$10\% for NMBE). Furthermore, the performance of LR and five deep learning models indicates higher predictability in heating than electricity since all of them achieved a lower CV-RMSE on predictions of heating than electricity. The higher predictability in heating load is because the building utilizes adaptive heating, which is driven by the difference between indoor and outdoor temperatures.

\begin{table}[!ht]
\centering
\caption{The prediction accuracy of point forecasts for different models on the test set. For both CV-RMSE and NMBE, a closer value to zero represents better prediction accuracy.}
\label{tab:point_forecast_result}
\begin{tabular}{@{}lrrcrr@{}}
\toprule
\multirow{2}{*}{Model} & \multicolumn{2}{c}{Electricity} & \phantom{} & \multicolumn{2}{c}{Heating} \\ \cmidrule(lr){2-3} \cmidrule(l){5-6} 
 & \begin{tabular}[c]{@{}r@{}}\textit{CV-RMSE}\\ \textit{(\%)}\end{tabular} & \begin{tabular}[c]{@{}r@{}}\textit{NMBE}\\ \textit{(\%)}\end{tabular} &  & \begin{tabular}[c]{@{}r@{}}\textit{CV-RMSE}\\ \textit{(\%)}\end{tabular} & \begin{tabular}[c]{@{}r@{}}\textit{NMBE}\\ \textit{(\%)}\end{tabular} \\ \midrule
SN-1 & 20.13 & $-$0.04 &  & 20.31 & $-$0.04 \\
SN-24 & 45.12 & $-$0.51 &  & 24.43 & $-$1.08 \\
LR & 14.65 & $-$0.49 &  & 11.55 & 0.45 \\
LSTM & 14.31 & $-$1.12 &  & 8.65 & 2.58 \\
TCN & \textbf{13.90} & 0.69 &  & \textbf{8.16} & 1.44 \\
Transformer & 14.93 & $-$2.59 &  & 10.04 & $-$2.26 \\
N-HiTS & 14.27 & $-$0.65 &  & 8.67 & 2.61 \\
TFT & 14.60 & $-$2.66 &  & 8.18 & 1.45 \\ \bottomrule
\end{tabular}
\end{table}

In contrast to the dominance of the TCN model in point forecast, none of the five deep learning models dominates the probabilistic forecast. For predicting the electricity, the TCN model performed best to capture the central tendency as it achieved the lowest $\rho$-risk at the 0.5th quantile ($\rho$-risk$(0.5)=0.0741$ as in Table~\ref{tab:quantile_forecast_result}). The LSTM model, on the other hand, performed best to capture the upper end of the distribution of the electricity ($\rho$-risk$(0.9)=0.0470$) and might be useful for predicting extreme values or identifying outliers. For predicting the heating, the TFT model performed best for both predicting median value ($\rho$-risk$(0.5)=0.0454$) and capturing the upper end of the distribution ($\rho$-risk$(0.9)=0.0231$).

\begin{table}[!ht]
\centering
\caption{The $\rho$-risk at 0.5th and 0.9th quantiles of probabilistic forecasts for different models on the test set. For each metric, lower values represent better performance.}
\label{tab:quantile_forecast_result}
\begin{tabular}{@{}lrrcrr@{}}
\toprule
\multirow{2}{*}{Model} & \multicolumn{2}{c}{Electricity} & \phantom{} & \multicolumn{2}{c}{Heating} \\ \cmidrule(lr){2-3} \cmidrule(l){5-6} 
 & $\mathit{\rho}$\textit{-risk(0.5)} & $\mathit{\rho}$\textit{-risk(0.9)} &  & $\mathit{\rho}$\textit{-risk(0.5)} & $\mathit{\rho}$\textit{-risk(0.9)} \\ \midrule
LSTM & 0.0784 & \textbf{0.0470} &  & 0.0462 & 0.0234 \\
TCN & \textbf{0.0741} & 0.0480 &  & 0.0468 & 0.0237 \\
Transformer & 0.0903 & 0.0483 &  & 0.0527 & 0.0266 \\
N-HiTS & 0.0833 & 0.0476 &  & 0.0678 & 0.0318 \\
TFT & 0.0812 & 0.0526 &  & \textbf{0.0454} & \textbf{0.0231} \\ \bottomrule
\end{tabular}
\end{table}

The probabilistic forecasts also show that heating is more predictable than electricity. When predicting heating rather than electricity, all models obtained decreased $\rho$-risk at 0.5th quantile. Meanwhile, the uncertainties in electricity consumption are greater than those in heating load since these models achieved a higher $\rho$-risk at 0.9th quantile when predicting electricity than heating. Nevertheless, The uncertainty in predicting electricity also indicates that, on the one hand, it is advantageous to improve confidence by optimizing electricity use while still assuring the regular functionality of a building. On the other hand, for more accurate forecasting, additional operational model-related features that impact electricity consumption should be included.

\subsection{Qualitative Analysis}

\begin{figure}[!ht]
\includegraphics[width=\columnwidth]{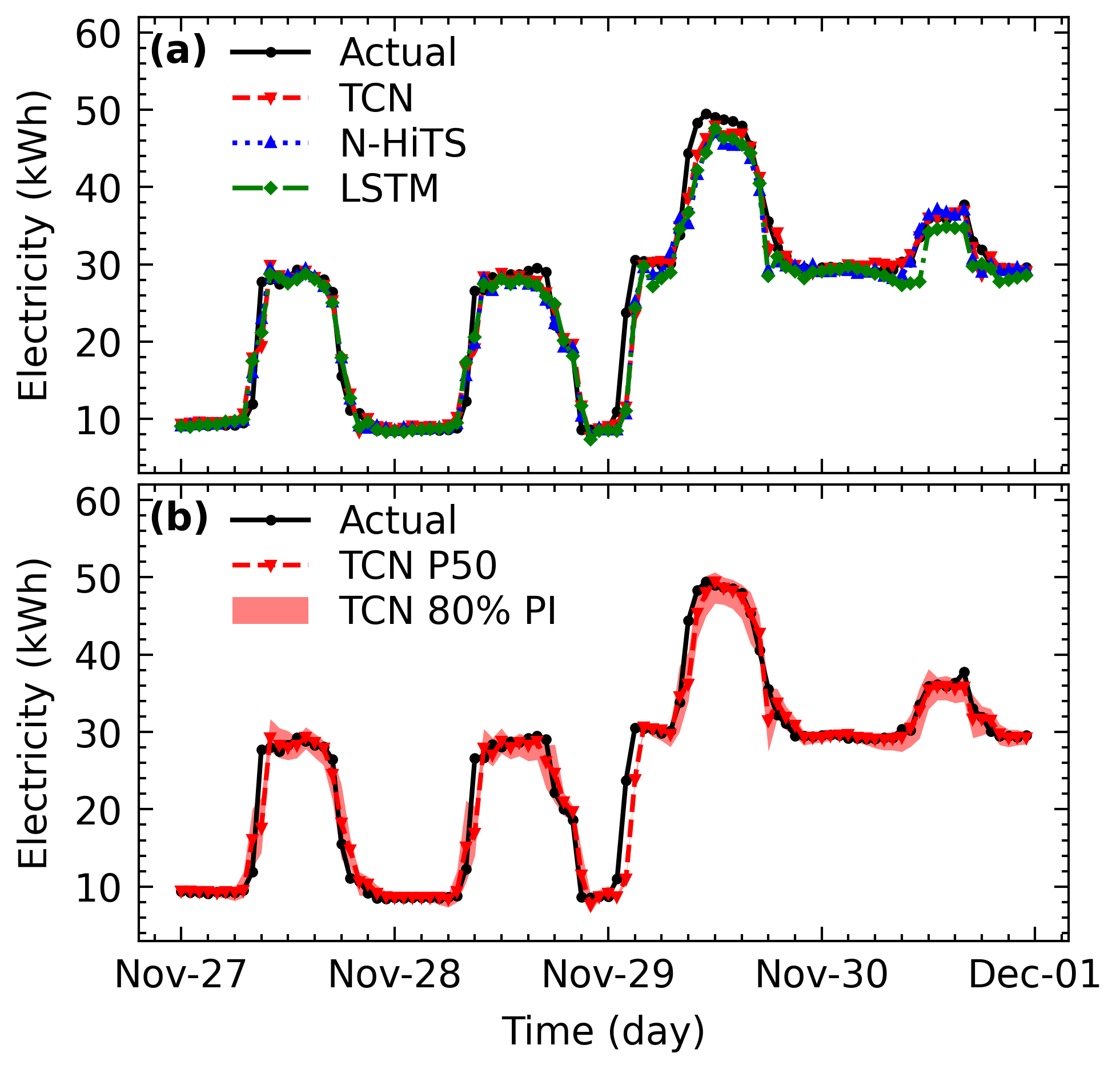}
\centering
\caption{The actual and predicted hourly electricity consumption from November 27 to November 30, 2019. (\textbf{a}) Point forecasts of the best three models and (\textbf{b}) probabilistic forecast of the TCN model. The predicted median is P50, and the 80\% prediction interval (PI) is from 0.1th to 0.9th quantile.}\label{fig:museum_electricity}
\end{figure}

Previous quantitative analysis indicates that heating load is more predictable than electricity consumption. The lower predictability was partly due to changes in the operating mode of the City Museum on some days in November and December 2019. Fig.~\ref{fig:museum_electricity}{a} shows such a change. During the two days, from November 29 to November 30, the hourly energy consumption in the nighttime was even higher than in the daytime of the previous days.

The changes in operating mode degrade the prediction accuracy of models during these days. On November 29 (the first day when the operating mode started to change), the predicted value has a certain lag (see Fig.~\ref{fig:museum_electricity}{a}). As shown in Fig.~\ref{fig:museum_electricity}{b}, the 80\% prediction interval (from 0.1th quantile to 0.9th quantile) during the daytime of the two days was relatively higher than during the daytime of the days before the operating mode changed.

\begin{figure}[!ht]
\includegraphics[width=\columnwidth]{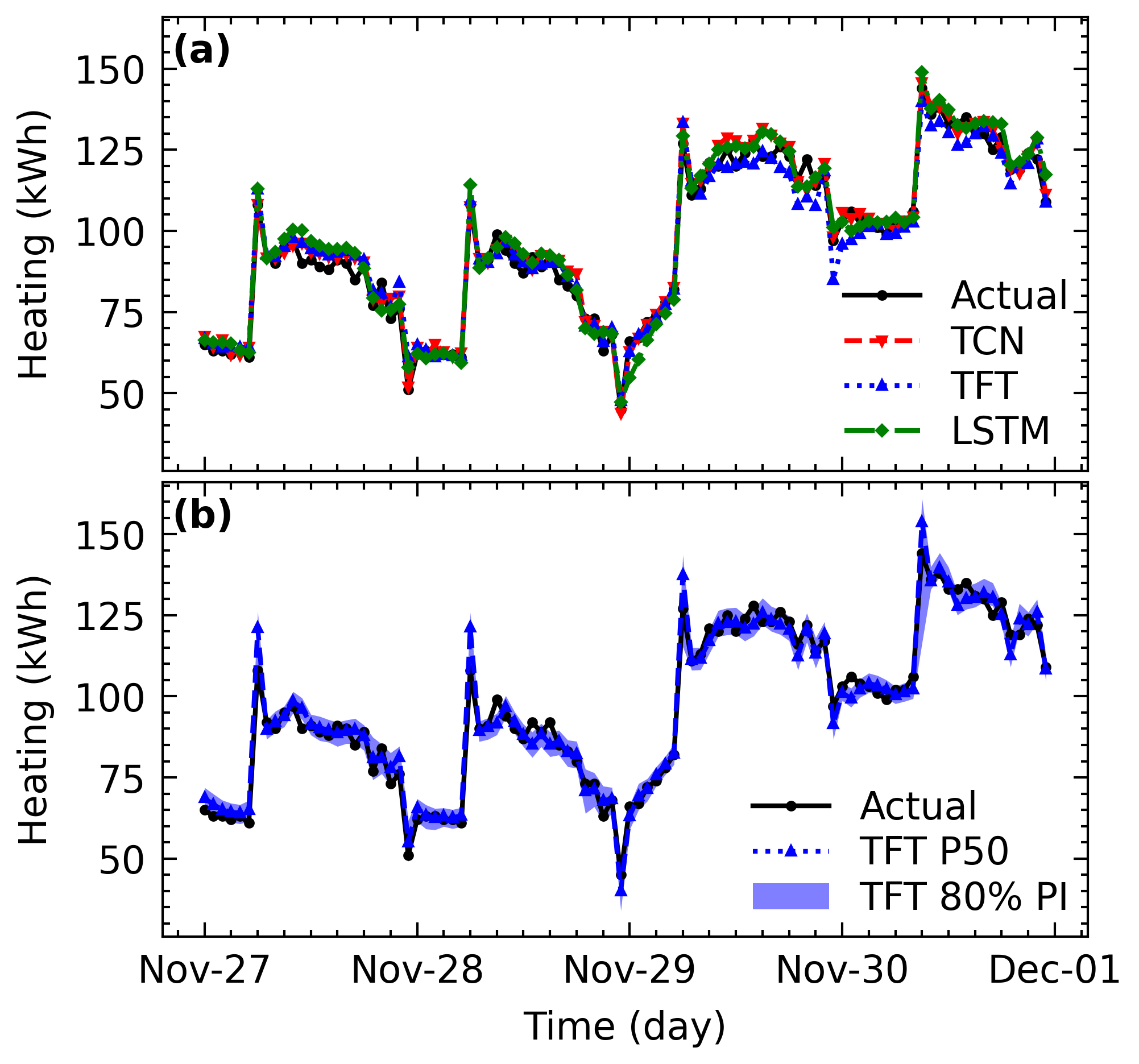}
\centering
\caption{The actual and predicted hourly heating load from November 27 to November 30, 2019. (\textbf{a}) Point forecasts of the best three models and (\textbf{b}) probabilistic forecast of the TFT model.}\label{fig:museum_heating}
\end{figure}

The higher predictability of heating load is attributed to strong influencing factors like dry-bulb temperature being involved in making predictions. In addition, the heating load is less affected by the change in operating mode. As shown in Fig.~\ref{fig:museum_heating}{a}, even on November 29 and 30, the two days when the operating mode changed, the best three models still made good predictions. Similarly, the uncertainty in predictions was greater during the daytime than during the nighttime (see Fig.~\ref{fig:museum_heating}{b}). A possible explanation for the higher uncertainty during daytime is that there is more heat exchange between the indoor and outdoor environments when more people are entering and exiting the building.

\section{Conclusion}
\label{sec:conclusion}

This study has presented a solution for integrating deep learning with digital twins to provide a more comprehensive understanding of building energy systems. Ontology was adopted for creating parametric digital twins of building energy systems to ensure a consistent data representation across various domains. Deep learning methods were applied to analyze the data collected by digital twins to identify patterns and seek the potential for saving energy. The results obtained from a case study in one public historic building in Norrköping, Sweden, have shown that deep learning methods, such as TCN, LSTM, and TFT, exhibit strong capabilities in capturing tendency and uncertainty in building energy consumption.

The solution provides facility managers with a better insight into building energy use. Thus, facility managers can proactively optimize energy systems to avoid unnecessary energy use. In the long run, this could result in cost savings, increased human comfort, and a more sustainable built environment.

\section*{Acknowledgment}
The authors thank Johan Bj\"orhn and his colleagues at Norrevo Fastigheter AB in Norrk\"oping for providing access to the City Museum and offering the historical energy consumption data. The Swedish Meteorological and Hydrological Institute is acknowledged for providing the weather data.

% The preferred spelling of the word ``acknowledgment'' in America is without 
% an ``e'' after the ``g''. Avoid the stilted expression ``one of us (R. B. 
% G.) thanks $\ldots$''. Instead, try ``R. B. G. thanks$\ldots$''. Put sponsor 
% acknowledgments in the unnumbered footnote on the first page.

% \section*{References}
\bibliographystyle{IEEEtran}
\bibliography{references}

\end{document}